\documentclass[runningheads]{llncs}

\usepackage{eccv}

\usepackage{eccvabbrv}

\usepackage{graphicx}
\usepackage{booktabs}

\usepackage{wrapfig}

\usepackage{multirow}
\usepackage{amssymb}
\usepackage{pifont}

\usepackage{footnote}

\newcommand{\cmark}{\ding{51}}%
\newcommand{\xmark}{\ding{55}}%

\usepackage[accsupp]{axessibility}  

\usepackage{hyperref}

\usepackage{orcidlink}

\definecolor{dartmouthgreen}{rgb}{0.05, 0.5, 0.06}
\definecolor{denim}{rgb}{0.08, 0.38, 0.74}
\definecolor{brightmaroon}{rgb}{0.76, 0.13, 0.28}

\begin{document}

\title{ViGoR: Improving Visual Grounding of Large Vision Language Models with Fine-Grained Reward Modeling} 



\author{Siming Yan\thanks{Equal contribution.}\textsuperscript{$\dagger$}\textsuperscript{1}
\hspace{0.15in}
Min Bai\textsuperscript{$\star$}\textsuperscript{2} 
\hspace{0.15in}
Weifeng Chen\textsuperscript{2} 
\hspace{0.15in} 
Xiong Zhou\textsuperscript{2} 
\hspace{0.15in} \\
Qixing Huang\textsuperscript{1} 
\hspace{0.15in}
Li Erran Li\textsuperscript{2}
\vspace{4pt}
}


\authorrunning{S.~Yan et al.}
\titlerunning{ViGoR}

\institute{\textsuperscript{1}The University of Texas at Austin \hspace{0.3in} \textsuperscript{2}AWS AI}

\maketitle

\begin{abstract}
By combining natural language understanding, generation capabilities, and breadth of knowledge of large language models with image perception, recent large vision language models (LVLMs) have shown unprecedented visual reasoning capabilities. However, the generated text often suffers from inaccurate grounding in the visual input, resulting in errors such as hallucination of nonexistent scene elements, missing significant parts of the scene, and inferring incorrect attributes of and relationships between objects.  
To address these issues, we introduce a novel framework, \textbf{ViGoR} (\textbf{Vi}sual \textbf{G}r\textbf{o}unding Through Fine-Grained \textbf{R}eward Modeling) that utilizes fine-grained reward modeling to significantly enhance the visual grounding of LVLMs over pre-trained baselines. This improvement is efficiently achieved using much cheaper human evaluations instead of full supervisions, as well as automated methods. 
We show the effectiveness of our approach through a variety of evaluation methods and benchmarks. 
Additionally, we released our human annotation (\href{https://github.com/amazon-science/vigor}{https://github.com/amazon-science/vigor}) comprising 15,440 images and generated text pairs with fine-grained evaluations to contribute to related research in the community.
\end{abstract} 
\footnotetext{Work done when interning at AWS AI. Contact email: siming@cs.utexas.edu.} 
\section{Introduction}
\label{sec:intro}
\begin{figure}
    \centering
    \includegraphics[width=\linewidth]{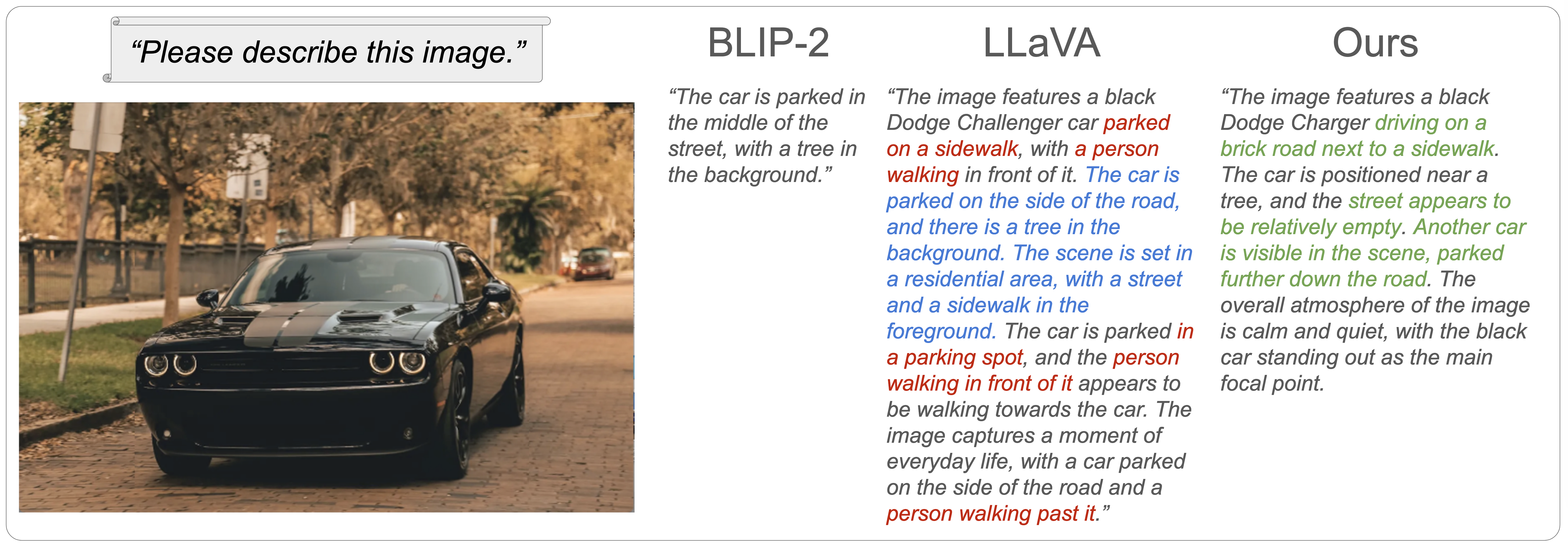}
    \caption{An illustration of inaccurate visual grounding in Large Vision Language Models.
    BLIP-2~\cite{li2023blip} failed to provide detailed image descriptions. LLaVA~\cite{liu2023llava}'s more detailed description contains both \textcolor{denim}{correct sentences} as well as \textcolor{brightmaroon}{sentences with hallucinations and inaccurate inference}. In contrast, our model preserves the logical reasoning and creativity of LVLMs while exhibiting \textcolor{dartmouthgreen}{significantly better accuracy and detail}.}
    \label{fig:teaser}
    \vspace{-15pt}
\end{figure}

Large language models (LLMs) have garnered intense interest throughout the research and academic communities. Typically, these models are pre-trained with tremendous amounts of automatically aggregated text data followed by further fine-tuning with specific user examples or evaluation feedback. This process enables the models to follow human-provided prompts and retrieve useful relevant information or solve logical problems. 
Recently, numerous techniques~\cite{li2023blip, dai2023instructblip, zhu2023minigpt, liu2023llava, ye2023mplug, li2023fine, yu2023reformulating} have enhanced these breakthroughs with the ability to understand visual information by further integrating image features into the prompt encoding process. 
Although these works have successfully aligned image features into the large language model domain, they still exhibit significant problems.
While a strong contextual grounding of language-only models can be learned from the enormous corpus of text data, paired training data for multimodal language/vision models is more limited, while the complexity of the task is arguably higher as the model must align two disparate modalities. Large, automatically compiled datasets, such as the LAION-5B dataset~\cite{schuhmann2022laion}, tend to feature only simple images with very short text descriptions. Training with such data often leads large vision language models to fail in capturing the essential details of the image, returning only a short and coarse description (see BLIP-2~\cite{li2023blip} output in Figure \ref{fig:teaser}). Moreover, techniques such as InstructBLIP~\cite{dai2023instructblip} and BLIP~\cite{li2023blip} primarily rely on high-quality paired language/image datasets (e.g. VQAv2~\cite{goyal2017making}, VizWiz~\cite{gurari2018vizwiz}, TextCaps~\cite{sidorov2020textcaps}, etc.). However, these datasets are expensive to collect and challenging to adapt for broader coverage due to the need for manual text annotation. On the other hand, initiatives such as LLaVA \cite{liu2023llava} train on perception and simple caption datasets in conjunction with the reasoning capabilities of LLMs to semi-automatically generate synthetic conversational ground truth. Unfortunately, such outputs can also contain non-factual statements suffering from hallucinations, omissions, and inaccuracies in attribute or relational descriptions, as they are generated by text-only models based on sparse information about the actual image. Hence, the resulting trained model is still not ideal (see LLaVA's output in Figure \ref{fig:teaser}).



Instead, we propose a novel and generally applicable framework using fine-grained reward modeling. It efficiently and substantially enhances the visual grounding of LVLMs beyond pre-trained baselines such as LLaVA, while preserving their capability to generate extended and detailed descriptions. Given a pre-trained LVLM (e.g., LLaVA), we input a set of images with prompts and generate multiple text outputs. Human annotators are asked to evaluate each image-text pair. As seen in the LLaVA output in Figure \ref{fig:teaser}, a lengthy description generated by a competent LVLM can contain both correct and incorrect sentences. Attempting to assign a single holistic score is ambiguous for the annotator, as well as a subsequently trained reward model. In our design, annotators assign fine-grained, per-sentence evaluation scores. This results in a newly compiled dataset comprising image-text-evaluation trios. 
We train a reward model to also predict dense reward scores and use it to fine-tune the pre-trained LVLM (see Section ~\ref{sec:method}). This fine-tuning process markedly improves the model's visual grounding capabilities with just 16K data samples, demonstrating the method's efficiency.

To further improve the performance of our system at negligible cost, we also develop a reward model scheme based on automatic methods \textit{without additional human effort}, which is proven to be highly effective in improving the visual grounding capabilities of LVLMs.

Finally, we amalgamated the strengths of both reward models to develop a complete solution, which we refer to as \textbf{ViGoR} (\textbf{Vi}sual \textbf{G}r\textbf{o}unding Through Fine-Grained \textbf{R}eward Modeling)
throughout the remainder of the paper.
 To assess the efficacy of ViGoR, we evaluate it against two commonly used benchmarks, POPE~\cite{Li-hallucination-2023} and MME~\cite{fu2023mme}, where it demonstrated significant improvements over the baseline models. 
 Finally, we compare the performance benefit of our approach to baselines for a variety of tasks that require accurate visual grounding. In summary, we make the following three major contributions.

\begin{itemize}

   \item We introduce a novel framework that incorporates fine-grained reward modeling with easily implemented rejection sampling, substantially enhancing the visual grounding of LVLMs.

   \item We develop reward models that require little human effort while leveraging the impressive advances in powerful and robust visual perception models, demonstrating marked improvements in visual grounding efficiency.

    \item We create and release the human evaluation dataset comprising 15.4K pairs of images and generated results, as well as the fine-grained assessments of the latter by our annotators. 
\end{itemize}
\vspace{-10pt}
\section{Related Work}
\label{sec:related}

\paragraph{Large Vision Language Models.}

Recent advances in LVLMs have been remarkable, such as the integration of large language models (LLM) such as GPT~\cite{brown2020language, radford2019language}, PaLM~\cite{chowdhery2022palm}, BLOOM~\cite{workshop2022bloom}, LLaMA~\cite{touvron2023llama}, and Vicuna~\cite{vicuna2023}. Flamingo~\cite{alayrac2022flamingo} and its open source counterpart OpenFlamingo~\cite{awadalla2023openflamingo}, along with IDEFICS, have been pivotal in integrating LLMs with vision-language pretraining, using techniques like gated cross-attention dense blocks. PaLI's~\cite{chen2023pali} research on the scaling of vision and language (V$\&$L) components across various tasks has been instrumental. As well, PaLM-E's~\cite{driess2023palme} extension of LLM to the embodied domain, and BLIP-2's~\cite{li2023blip} introduction of the Querying Transformer (Q-former) to align image and language encoders mark significant progress. InstructBLIP~\cite{dai2023instructblip} further enhances this approach. Otter's~\cite{li2023otter} improves OpenFlamingo’s instruction-following capabilities and MiniGPT-4's~\cite{zhu2023minigpt} recommendation for a single project layer to align visual and linguistic models demonstrate its efficiency and capability. mPLUG-Owl~\cite{ye2023mplug} aligns visual characteristics before fine-tuning the language model using LoRA~\cite{hu2022lora}. LLaVA directly injects visual tokens into a pre-trained LLM and finetunes the model with synthetic conversations generated by GPT-4 using metadata and short captions as input. 
While these techniques primarily focus on architectural innovation for aligning image features to the feature space of LLMs or leveraging extensive image-text data for instruction tuning, our work introduces a distinct and novel general framework designed to enhance the visual grounding capability of any LVLM.

\paragraph{Visual Perception Models.}

Recent advances in visual~\cite{radford2021clip,li2022grounded,caron2021emerging,liu2023grounding} and 3D~\cite{yang2020extreme, zhuang2021unsupervised, yan2022implicit, yan2023multi, yan2023implicit, IAE:SSR:2022, YanYMHVH21, yan2021hpnet, zhuang2019self, yan20233d, yan2019recurrentfeedbackimprovesfeedforward, dgcnn:2019:tog,xie2020pointcontrast, Yang2018a} perception models have demonstrated remarkable proficiency in handling a variety of tasks in open world scenes. In particular, CLIP~\cite{radford2021clip} has exhibited a robust capability for zero-shot classification. By reformulating object detection as a grounding problem, GLIP~\cite{li2022grounded} has achieved semantic alignment at both the phrase and region levels, leading to impressive open-set detection performance. GroundingDINO~\cite{liu2023grounding} represents another significant step, grounding the state-of-the-art transformer-based object detector DINO~\cite{caron2021emerging} with language pretraining for open-set generalization. This model demonstrates a strong ability to discern whether elements are present in a scene and accurately detect object counts. 

However, such advanced visual grounding abilities have not yet been fully realized in current LVLMs. To bridge this gap, we devise an automatic method for building a reward model using these vision perception models, and distill their strong visual grounding capabilities directly into LVLMs.

\vspace{-0.2cm}

\paragraph{Reward Modeling.} Recent progress in training Large Language Models (LLMs) has increasingly emphasized the importance of reward modeling. This approach often incorporates human feedback and reinforcement learning optimization strategies, such as Proximal Policy Optimization (PPO). This approach is crucial in refining the accuracy and contextual relevance of model outputs. For example, Askell et al. ~\cite{askell2021general, instructGPT2022} highlighted the potential of using human feedback in the training of general language assistants, emphasizing the importance of aligning model responses with human standards and values. Recently, LLaMA-2~\cite{llama2} introduced a novel rejection sampling strategy within reward modeling, claiming that it improves the generation of contextually appropriate high-quality responses.

However, in the realm of Large Vision Language Models (LVLMs), the application of reward modeling remains underexplored, with most existing work focusing predominantly on instruction tuning~\cite{liu2023llava, zhu2023minigpt}. One exception is LLaVA-RLHF which adapts the Reinforcement Learning from Human Feedback (RLHF) from the text domain to the task of vision-language alignment, where human annotators are asked to compare two responses and pin-point the more hallucinated one~\cite{2023llavarlhf}. Inspired by LLaMA-2, we combine reward modeling with rejection sampling in the LVLM training framework. While LLaVA-RLHF uses a reward model that produces sparse signals, we leverage fine-grained reward models to improve the visual grounding capabilities of LVLMs, leading to more accurate and contextually relevant output in vision-language tasks.

\begin{figure*}[t]
    \centering
    \includegraphics[width=1.0\textwidth]{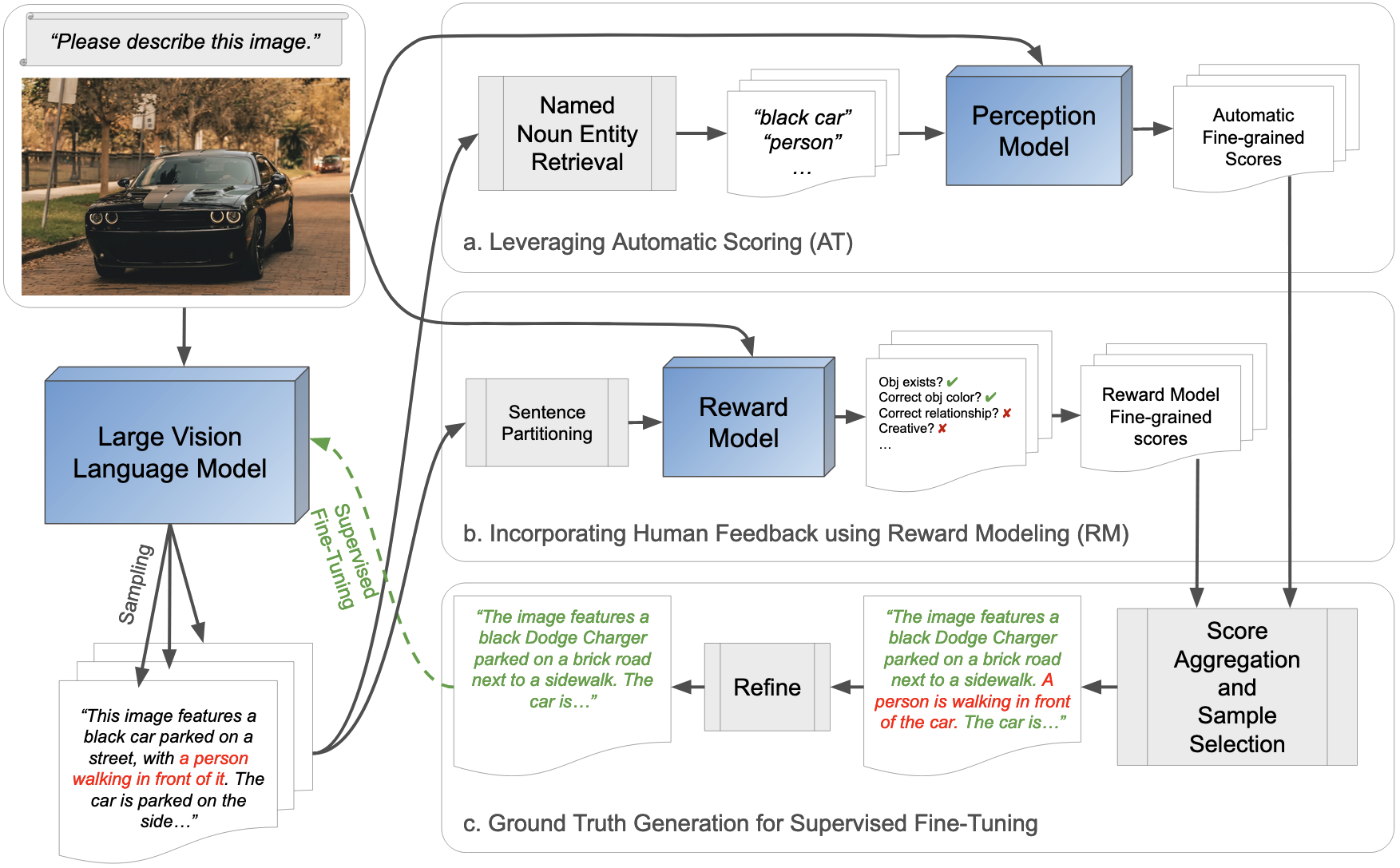}
    \caption{\textbf{Overview of our model training framework.} Starting with an input image and a prompt to generate a detailed caption, we sample a number of responses from the LVLM. These responses are passed through two fine-grained reward signal generation branches (\textit{a.} by leveraging state-of-the-art generalizable perception models and \textit{b.} by using an LVLM-based reward model trained using annotator feedback). Finally, we \textit{c.} combine the fine-grained assessment signals from both sources into a single reward score, and select the best sampled description. Finally, we use heuristics and byproducts from the automated scoring system to further refine this sample, and use it for supervised fine-tuning of the LVLM.}
    \label{fig:framework}
    \vspace{-15pt}
\end{figure*}


\section{ViGoR: Visual Grounding Improvement Framework}
\label{sec:method}

Our primary goal is to increase visual grounding and reduce hallucinations while keeping the strong intuitive reasoning and creative thought process of pre-trained LLMs and LVLMs. While it has been shown that high quality human annotations for supervised fine-tuning is a straightforward approach for significantly improving LVLMs, it is cost-prohibitive for many application scenarios. As such, we wish to efficiently leverage annotator time and the latest advances in direct visual perception models such as powerful open-set object detectors. 

We construct a system to fine-tune a base LVLM with rejection sampling, similar to LLaMA-2~\cite{llama2}, allowing the model to improve through using intelligent ranking of its own sampled output. This requires a robust and perceptive scoring system for the generated text. In our work, we use two complementary solutions in parallel. In particular, we train a reward model to incorporate human annotator assessments of text outputs from the LVLM, and provide positive and negative assessments of the LVLM during training time with unlabeled examples. As well, we leverage an open-set object detector and heuristics to verify the existence or absence in the image of the named noun entities extracted from the generated descriptions. Finally, we combine these signals into a single reward score, and use it to select the best description among the initial samples. This sample undergoes additional refinement and is used for the supervised fine-tuning of the LVLM. We refer the reader to Figure \ref{fig:framework} for a visualization.




\subsection{Reward Modeling via Fine-Grained Human Feedback}

\paragraph{Human Preference Data Collection.} We design a system to incorporate human judgment and preference --- considered the most reliable ground truth signal --- into our model training. First, we select a pretrained LVLM's checkpoint state, and create image / caption pairs from the model using a nonzero temperature to balance factual rigor with creativity. Unlike other approaches that ask annotators to provide relatively sparse judgments for the LLM/LVLM's generated results, we ask crowd-workers to provide fine-grained feedback at the sentence level. For each sentence containing errors, the annotator selects the nature of the inaccuracy from a predefined list including object hallucinations, type of attribute error, incorrect relationships between multiple objects, error in the mentioned location or unreasonable conjectures about the image. Furthermore, the annotator provides a judgment about the creativity of the sentence. The creativity means if the sentence gives a thoughtful and reasonable interpretation or extrapolation of the image.
Finally, the annotator provides a holistic assessment of the overall description's level of detail and identifies the missing elements in the scene. 
These requirements encapsulate our overall goal: to enhance the LVLM’s visual grounding across the entire image while maintaining the insight and creativity inherent in the pre-trained language decoder. 

\paragraph{Reward Model Training.}\label{rm_training} Using the collected annotations, we fine-tune a dedicated LVLM as the reward model on the annotations using instruction tuning to judge the base LVLM's generated results during training time. The reward model is trained to output a sequence of text tokens which encodes the various scores given the underlying image and the LVLM's output as the input. While existing work~\cite{2023llavarlhf} generates a single holistic score for the entire text output, this process can be ambiguous when the description contains both correct and incorrect components (see Figure \ref{fig:teaser}). Instead, we train the model to produce sentence-level evaluations. This fine-grained approach reduces ambiguity, and increases the detailed visual grounding of the reward model. To provide the necessary context for scoring each sentence, we prepend the sentence with all preceding generated text and explicitly ask the reward model to score the last sentence (which is also the target sentence) in the given passage. A typical prompt is ``\textit{Assess the accuracy of the last sentence in the following description of this image, and return a single number.}" Due to the fine-grained feedback provided by the annotators, these prompts result in either a positive response (when no errors are found in the sentence), or a detailed negative response (where one of several error types are found in the sentence). 
More details are found in the Supplementary Materials. 

As we show in the ablation studies, compared with holistic-based method, this fine-grained method significantly improves the reward model's capabilities to guide the fine-tuning of the LVLM. Furthermore, we see that the reward model trained with fine-grained feedback can better understand the link between errors in the descriptions and the image, resulting in superior fine-tuning performance.

\subsection{Reward Modeling with Automatic Methods}

While the preference annotations directly encapsulate human preferences and are cheaper than supervised fine-tuning annotations, the human effort is still non-negligible. This can limit the scale of potential datasets and subsequently the visual discerning capabilities of the learned reward model due to overfitting. To further improve the cost efficiency of the overall system, we leverage the advances in discriminative vision perception models to automatically score the grounding and fidelity of text generated by a LVLM on large quantities of unlabeled images. However, these discriminative models generally have structured input (such as images and semantic classes) and structured outputs (such as bounding boxes). On the other hand, LVLMs operate with unconstrained and unstructured input and output (free-form text). This gap prevents directly providing the LVLM's output to the discriminative models for scoring. As such, we carefully design the scoring system shown in part \textit{a} of Figure \ref{fig:framework}. 

Starting with a caption generated by the LVLM from an image, we identify the individual nouns mentioned using standard named entity recognition with NLTK \cite{BirdKleinLoper09}. Next, we prompt an open-set object detector with these nouns to verify the existence of the objects in the image. A correct identification is rewarded with a positive score, while hallucinated objects incur a penalty. 
We use Grounding DINO~\cite{liu2023grounding} for its strong performance across a wide variety of image domains. However, we note that while these detection models are adept at identifying the existence of objects, their ability to detect other types of errors is limited (e.g. object attributes and relationships). This limitation underscores the continued necessity for human-preference-based reward modeling. 


\subsection{Reward Score and Rejection Sampling} 

As both our reward model-based and automated methods provide fine-grained reward scores, we must design a strategy to combine them into a single score to enable rejection sampling at the description level. Note that this process embodies the same ambiguity faced by annotators in existing work, where they are asked to analyze complete text output in detail and rank them based on \textit{holistic} preference. However, our solution combines the detailed analysis in a more principled and consistent fashion.

The sample with the best score is used as ground truth for supervised fine-tuning. As our focus is to reduce generation of erroneous descriptions, we use the signals that indicate errors as the primary selector. For each sample, we aggregate all negative signals from the two streams of description evaluation by normalizing their values with their respective variances and linearly combine them. We select the sample with the smallest penalty score as the best candidate, and use the positive scores as a tiebreaker when necessary. 


\subsection{Refinement Module}
We design a simple refinement module to further polish the selected candidate description to use as a regression target (see part \textit{c} of Figure \ref{fig:framework}). For each sentence within the description, we assess the presence of noun phrases deemed nonexistent by the reward modeling module. Should a sentence contain any such noun phrases, we eliminate that sentence entirely from the description. For instance, consider the description:\textit{``The image features a black Dodge Charger parked on a brick road next to a sidewalk. A person is walking in front of the car."} If the reward modeling module identifies \textit{``person"} as a non-existent noun phrase, the sentence \textit{``A person is walking in front of the car."} is removed. This refinement process
effectively reduces the hallucination problem by removing inaccurate elements, thereby leading to notable improvements in the model's performance.

\subsection{Model Training}
The resulting description from the refinement module is used as ground truth in supervised fine-tuning with the standard autoregressive objective as in the original LLaVA \cite{liu2023llava}. As will be demonstrated by the ablation studies, two signal sources are complementary and provide better results than either one alone. This overall process is visualized in Figure \ref{fig:framework}.

\section{Experiments and Results}
\label{sec:results}

We compare the effectiveness of our approach with competitive baselines, and delve into the contributions from each component in our design. As well, we provide visualizations to qualitatively compare the output of fine-tuned model with that of its initial state. In the following, ``ViGoR-AT" refers to our approach utilizing only reward modeling with automatic methods. ``ViGoR-RM" represents our method employing reward modeling with fine-grained human feedback. ``ViGoR-All" and ``ViGoR" denote our method with the combination of the reward modeling with automatic methods and fine-grained human feedback. 
\subsection{Testing Framework}

We use the recently proposed LLaVA \cite{liu2023llava} as the base model to demonstrate the effectiveness of our fine-tuning strategies without modifications to its architecture. In particular, we select the variant of the LLaVA model with the pre-trained and frozen ViT-L / 14 @ 224px CLIP image encoder \cite{radford2021clip} and the pre-trained Vicuna v1.3 \cite{vicuna2023} with 7B parameters as the language model. Our method is not specific to any particular LVLM testing architecture or configuration. We use this configuration for both our base LVLM model and our learned reward model. 

With computation efficiency in mind, we select the smallest variant of the Grounding DINO model with the Swin-T~\cite{liu2021swin} backbone with the official checkpoint and the default box and text thresholds (0.25) for the open-set object detection task. Our proposed method can likely directly benefit from the larger and more computationally expensive variants or future advances in these models. 

\subsection{LVLM Model Fine-tuning}

To support our experiment, we used 25,574 images from the ADE20K dataset \cite{zhou2017ade20k} (training split) as the basis, as they cover a wide variety of relatively complex scenes. To preserve the generalization of our method, we disregard the original annotations of the dataset. We elicit responses from our base LVLM through a set of question prompts (e.g. ``\textit{Please provide a detailed description of the given image.}"), and sample 5 outputs for each input image. Subsequently, these outputs are fed into our reward model to obtain a corresponding reward score for each description. In addition to keeping the highest-scoring description following the common practice in rejection sampling, we enhance this description further by applying a refinement module to achieve additional quality improvements.

To reduce computational complexity, we choose an \textit{offline} supervision generation strategy where we save the refined output as the ground truth for its respective image using the initial model state. However, we note that the rejection sampling process can take place \textit{online} by refining an evolving model state. We believe that this has the potential to further improve the algorithm's effectiveness, but requires significantly more computation. 

During the fine-tuning phase, we set the learning rate at $2\times10^{-5}$ and train the model over two epochs with a batch size of 32. The entire process is executed on eight 40G A100 GPUs, taking 7 hours in total to complete.

\subsection{Reward Model Training}

\paragraph{Human feedback collection.} To allow the reward model to be highly receptive to errors in the descriptions, we use a pre-trained checkpoint for LLaVA that is already fine-tuned with our automatic reward generation mechanism. Through hands-on experimentation, we observed that while the images from the ADE20K dataset exhibit excellent scene variability, the resulting cognitive load for human annotators is very high. As such, instead, we generate 15,440 detailed image captions using images selected from the somewhat simpler MS COCO~\cite{lin2014microsoft} dataset. We enlist the services of 15 professional annotators to assist us in creating the evaluation data set for the training of the reward model using the process described in Section~\ref{rm_training}. We provide carefully designed and detailed annotation instructions (available in our Supplementary Materials) to our annotators, along with extensive sample annotations. The process took approximately 3 weeks. 

\paragraph{Reward model.} We initialize the same model architecture as our base LVLM with the same weights used to generate the annotation samples as the starting state of our reward model. We train this model for 5 epochs on the dataset with 15,440 samples with a batch size of 32 and an initial learning rate of $2\times10^{-5}$. 

\begin{table}[t]
\footnotesize
    \centering
    \setlength\tabcolsep{5pt}
    \begin{tabular}{l|ccccccc|c}
        \toprule
        Method & HL & CA & AA & RA & RL & RS & DL & Average \\
        \midrule
        LLaVA & 3.28 & 3.27 & 3.28 & 3.27 & 3.27 & 3.27 & 3.01 & 3.24\\
        ViGoR-AT & 2.26 & 2.27 & 2.25 & 2.26 & 2.25 & 2.30 & 2.06 & 2.24 \\
        ViGoR-RM & 2.48 & 2.50 & 2.47 & 2.49 & 2.49 & 2.46 & 2.96 & 2.55 \\
        ViGoR-All & \textbf{1.97} & \textbf{1.96} & \textbf{1.99} & \textbf{1.98} & \textbf{1.99} & \textbf{1.97} & \textbf{1.97} & \textbf{1.97} \\
        \bottomrule
    \end{tabular}
    \caption{\footnotesize{\textbf{Detailed description generation evaluation.} Scores are the \textit{average preference rank} of the caption when compared within the 4 candidates in the table as judged by GPT-4Vision (ranging from 1 to 4; lower is better). \\
    HL: Does the description have \textbf{hallucinations}? \\
    CA: Does the description have good \textbf{counting accuracy} for objects? \\
    AA: Does the description assign \textbf{accurate attributes} to objects? \\ 
    RA: Does the description have \textbf{accurate relationships} between objects? \\ 
    RL: Is the description \textbf{relevant} to the image as a whole? \\
    RS: Does the description exhibit reasonable thought and \textbf{reasoning}? \\
    DL: Does the description encompass all \textbf{details} in the image? 
    }}
    \label{tab:gpt4v}
\vspace{-20pt}
\end{table}

\subsection{Quantitative Results}

The types of natural interactions that a user can have with a LVLM are essentially unlimited in variety. Since the focus of our work is on improving visual grounding, we primarily evaluate the benefits of our technique in three types of settings. First, we analyze the unstructured output of the LVLM when it is asked to generate a detailed and comprehensive caption for an image. This tests the model's ability to be perceptive to all regions and objects in the image. Second, we evaluate the ability of the model to form short sentences in response to targeted questions where some logical reasoning is needed. Lastly, we evaluate the model's performance on responding to highly specific questions for which a concise and unambiguous answer is required. This further tests the model's ability to use the input prompt to guide its visual understanding. We submit that this variety of tasks forms a comprehensive testing framework. 

\vspace{-15pt}

\subsubsection{Detailed Description Generation Evaluation.} 
\label{gpt4}

Examining the factuality of detailed descriptions generated by a LVLM based on an input image allows provides highly comprehensive insight into visual grounding capabilities. However, objective quantitative measurement of free-form responses has been a major challenge in the LLM and LVLM space due to the inherent ambiguities in defining quality. Furthermore, it is unclear how structured information can be extracted from generated text so that a score can be assigned using traditional objective functions. As demonstrated by \cite{helm2024}, the GPT family of models have very high levels of agreement with human annotators when asked to follow instructions to evaluate output of other models. This is likely due to the large quantities of proprietary data used for training. Therefore, we ask GPT-4Vision to rank the output of different LVLM candidates according to numerous criteria that signify the quality of the captions. This offers a detailed view of the visual foundation and text generation capabilities of the LVLMs under test. 


For detailed insight, we evaluate model output in terms of hallucinations, relevance, reasoning, level of detail, as well as accuracy in counting, attributes, and relationships between objects. As it is difficult to grade lengthy text responses against an absolute scale, we instead use preference ranking. GPT-4Vision is asked to consider several candidate responses relative to one another against the sample input image and text query, and provide a ranked list over the responses for each metric. The rankings are averaged over all samples to obtain an overall ranking score for each model, where lower is better. 

We evaluate using 300 randomly selected images from the COCO validation dataset and show the results in Table~\ref{tab:gpt4v} along with an overview of the definitions of the metrics. We select the four most relevant configurations for comparison: 1) the original LLaVA baseline model that we use as the initialization; 2) ViGoR-AT: the ViGoR method using only reward modeling with automatic methods; 3) ViGoR-RM: the ViGoR method using only the guidance from the reward model trained using the annotator feedback, and lastly 3) ViGoR-All: the complete model. As is clearly evident, our complete model consistently achieves the best ranking within the comparison across all metrics. Furthermore, we observe that either source of supervision signals significantly outperforms the original baseline model. Further details are provided in the Supplementary Materials.



\begin{wraptable}[12]{r}{0.5\textwidth}
    \vspace{-25pt}
    \centering
    \begin{tabular}{lcc}
    \toprule
         Method & \#Param & MMHal-Bench\\
        \midrule
        LLaVA & 7B & 1.3 \\
        LLaVA-RLHF & 7B & 1.4 \textcolor[rgb]{0,0.6,0}{(+0.1)} \\
        ViGoR (ours) & 7B & \textbf{1.6} \textcolor[rgb]{0,0.6,0}{(+0.3)} \\
        \bottomrule
    \end{tabular}
    \captionof{table}{\textbf{Short answer evaluation on MMHal-Bench benchmark.} For a fair comparison, we show the results of LLaVA-RLHF using solely RLHF optimization without supervised fine-tuning using additional data.}
    \label{tab:mmhal}
\end{wraptable}

\vspace{-5pt}
\subsubsection{Short Answer MMHal-Bench Evaluation.} 

Next, we showcase the performance of our system on the recently proposed MMHal-Bench \cite{2023llavarlhf}, which involves 96 questions of a variety of types to which short answers are often appropriate, such as ``\textit{How would you describe the weather in the image?}". The corresponding images are hand-selected from OpenImages. The evaluation process sends ground truth annotations and the model's responses to GPT-4, and requests GPT-4 to judge the descriptiveness and level of hallucinations. This is combined into a composite score (higher is better). 


We show the results in Table~\ref{tab:mmhal}, consisting of the average score as reported by the MMHal-Bench. We see that our approach results in a much larger gain in performance (+0.3) compared to the baseline established by LLaVA-7B, which is more substantial than the improvement achieved through the RLHF scheme proposed by LLaVA-RLHF~\cite{2023llavarlhf} without using supervised fine-tuning (+0.1). We further note that our particular implementation of the ViGoR framework only uses prompts to elicit detailed descriptions of the entire scene, and does not cover additional types of questions. Therefore, the encouraging results on MMHal-Bench suggests that the improvements generalize well to different types of queries. As the underlying images do not come from MS COCO or ADE20K (the data sources for our fine-tuning process), this further shows the generalization capability of our approach.

\vspace{-5pt}
\subsubsection{Short Answer Programmatic Benchmark Evaluation.}
Finally, we evaluate the models using benchmarks which consist of questions with concise and unambiguous one word answers such as ``\textit{Is there a red apple in the image?}" These benchmarks avoid the uncertainty and noise introduced from leveraging a third-party large language model for evaluation, and instead uses programmatic comparisons. Similar to MMHal-Bench, these benchmarks can measure both the visual grounding capabilities and the generalization of our method as the question types are outside the scope of the training data. We select the commonly used POPE~\cite{Li-hallucination-2023} and MME~\cite{fu2023mme} benchmarks.
POPE probes the LVLM with 3000 \textit{yes/no} questions targeting 500 images from COCO~\cite{lin2014microsoft} to evaluate the LVLM's ability to determine the existence of specific objects in a scene. MME extends the questions to cover aspects such as numerical count, position, and color of objects, as well as other vision tasks such as scene/landmark identification and OCR from data sources other than MS COCO and ADE20K.

\begin{table}[t]
    \footnotesize
    \centering
   \setlength\tabcolsep{4pt}
    \begin{tabular}{lccrrrrrrr}
        \toprule
     \multirow{2}{*}{Method} & \multirow{2}{*}{\#P} & \multirow{2}{*}{VQA} & \multicolumn{2}{c}{Random} & \multicolumn{2}{c}{Popular} & \multicolumn{2}{c}{Adversarial} & Avg F1\\
         & & & AC & PR & AC & PR & AC & PR & \\
        \midrule
        mPLUG-Owl & 7B & \xmark & 53.3 & 51.7 & 50.6 & 50.3 & 50.7 & 50.3 & 67.2 \\
        MiniGPT-4 & 7B & \cmark & 77.8 & 75.4 & 68.3 & 64.3 & 66.6 & 62.45 & 74.1 \\
        MM-GPT & 9B & \cmark & 50.0 & 50.0 & 50.0 & 50.0 & 50.0 & 50.0 & 70.1 \\
        InstructBLIP & 13B & \cmark & \textbf{88.7} & 85.0 & 81.4 & 75.1 & 74.4 & 67.7 & 83.7 \\
        \midrule 
        LLaVA & 7B & \xmark & 54.4 & 52.3 & 52.4 & 51.3 & 50.8 & 50.4  & 67.8 \\
        LLaVA-RLHF & 7B & \xmark & - & - & - & - & - & - & 78.2 \\
        LLaVA-RLHF & 7B & \cmark & - & - & - & - & - & - & 82.7 \\
        ViGoR (ours) & 7B & \xmark & 85.1 & \textbf{89.0} & \textbf{81.5} & \textbf{83.0} & \textbf{75.5} & \textbf{73.8} & \textbf{83.8} \\
        \bottomrule
    \end{tabular}
    \caption{\small{\textbf{Quantitative results on POPE benchmark.} POPE is comprised of three parts, each generated by different sampling strategies: Random, Popular, and Adversarial Sampling. We report the Accuracy(AC) and Precision(PR) for each part, alongside the average F1 score (Avg F1) across all three parts.}}
    \label{tab:pope}
    \vspace{-10pt}

\end{table}

\begin{table*}[t]
\scriptsize
    \centering
   \setlength\tabcolsep{0.3pt}
    \begin{tabular}{lcccrrrrrrrrrrrr}
        \toprule
     Method & \#Param & Data & VQA & EX & CT & PO & CO & PT & CE & SC & LM & AT & OC & Overall \\
        \midrule
        MiniGPT-4 & 7B & 5K & \cmark & 68.3 & 55.0 & 43.3 & 75.0 & 41.8 & 54.4 & 71.8 & 54.0 & 60.5 & 57.5 & 581.7 \\
        VPGTrans & 7B & 1.4M & \xmark  & 70.0 & 85.0 & 63.3 & 73.3 & 84.0 & 53.5 & 141.8 & 64.8 & 77.3 & 77.5 & 790.5\\
        InstructBLIP & 13B & 1.2M & \cmark & 185.0 & 143.3 & 66.7 & 153.3 & 123.8 & 101.2 & 153.0 & 79.8 & 134.3 & 72.5 & 1212.8 \\
        Cheetor & 7B & 500K & \cmark & 180.0 & 96.7 & 80.0 & 116.7 & 147.3 & 164.1 & 156.0 & 145.7 & 113.5 & 100.0 & 1299.9 \\
        Muffin & 13B & 267K & \cmark & 195.0 & 163.3 & 66.7 & 165.0 & 137.8 & 81.8 & 151.3 & 146.3 & 116.5 & 57.5 & 1281.0 \\
        \midrule
        LLaVA\footnotemark & 7B & 158K & \xmark & 158.3 & 83.3 & 51.7 & 85.0 & 94.2 & 85.0 & 145.8 & 125.8 & 74.3 & 57.5 & 960.1 \\
        ViGoR (ours) & 7B & 174K & \xmark & 180.0 & 143.3 & 83.3 & 100.0 & 128.2 & 121.2 & 150.5 & 130.8 & 127.0 & 145.0 & 1309.3 \\
        \textit{Improvement} & & & & {\textcolor[rgb]{0,0.6,0}{+21.7}} & {\textcolor[rgb]{0,0.6,0}{+60.0}} & {\textcolor[rgb]{0,0.6,0}{+31.6}} & {\textcolor[rgb]{0,0.6,0}{+15.0}} & {\textcolor[rgb]{0,0.6,0}{+34.0}} & {\textcolor[rgb]{0,0.6,0}{+36.2}} & {\textcolor[rgb]{0,0.6,0}{+4.7}} & {\textcolor[rgb]{0,0.6,0}{+5.0}} & {\textcolor[rgb]{0,0.6,0}{+52.7}} & {\textcolor[rgb]{0,0.6,0}{+87.5}} & {\textcolor[rgb]{0,0.6,0}{+349.2}}\\
        
        \bottomrule
    \end{tabular}
    \caption{\small{\textbf{Quantitative results on MME benchmark.} `VQA' indicates whether the model is fine-tuned using the VQA datasets. Abbreviations explained: `EX': Existence; `CT': Count; `PO': Position; `CO': Color; `PT': Poster; `CE': Celebrity; `SC': Scene; `LM': Landmark; `AT': Artwork; 'OC': OCR. For a fair comparison, we compare the Large Vision Language Model using the Vicuna backbone.}}
    \label{tab:mme}

\vspace{-20pt}
\end{table*}
\footnotetext{We report the number of LLaVA v1.0 reproduced by ourselves.}


As illustrated in Tables~\ref{tab:pope} and ~\ref{tab:mme}, our model shows consistent improvement across all categories evaluated in both datasets. Furthermore, while the competitive baseline methods~\cite{zhu2023minigpt, dai2023instructblip, yu2023reformulating, li2023fine} use VQA datasets containing ground truth for these types of straightforward questions, our method is able to achieve comparable or superior results \textbf{without fine-tuning on such resources}. This suggests that the visual grounding abilities learned by the model through our process is general. When the model learns to generate accurate \textit{comprehensive descriptions} for images, it is able to apply the capabilities to related \textit{reasoning} tasks, suggesting that the logical deduction capabilities of the underlying pre-trained LLM is retained and leveraged. Furthermore, our 7B parameter model shows on par performance with some models with nearly double the number of parameters (13B). This underscores the efficiency and robustness of our model in handling diverse visual-language tasks. As with the case of MMHal-Bench, the MME benchmark draws imagery from sources other than MS COCO to encompass posters, celebrity images, scenes, landmarks, artwork, and OCR. Our method (using ADE20K and MS COCO in the pipeline) still shows significant improvements in these categories, further proving its generalizability.

In Table~\ref{tab:pope}, we also compare our method with LLaVA-RLHF~\cite{2023llavarlhf}. The results indicate that our method outperforms LLaVA-RLHF, even when the latter is trained with additional VQA data. This observation further demonstrates the superior visual grounding capabilities of our model, attributed to our fine-grained reward modeling strategy.

\begin{figure}[t]
    \centering
    \includegraphics[width=\linewidth]{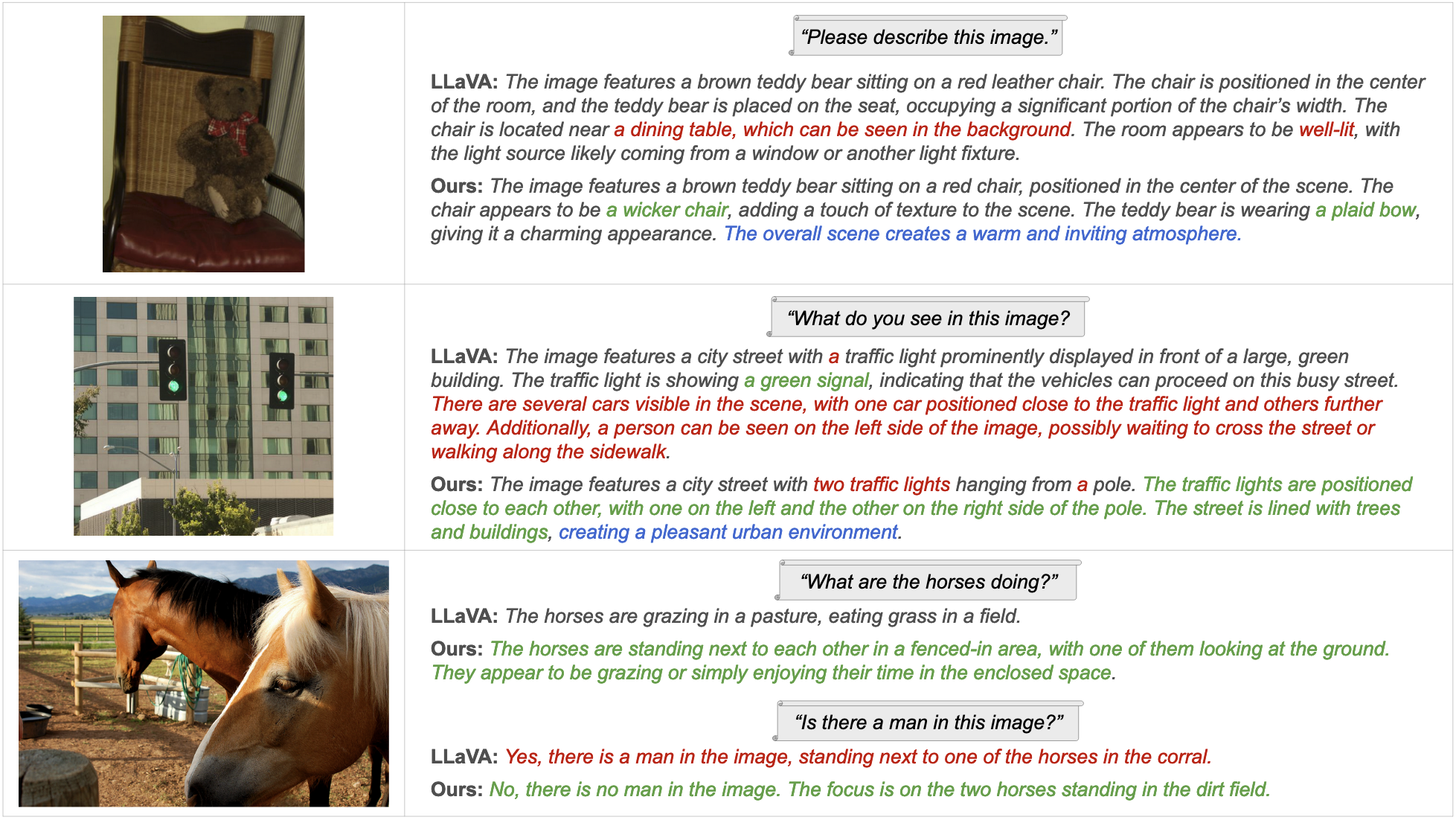}
    \caption{\textbf{Qualitative results.} We show examples of descriptions generated by our technique after using the fine-tuning scheme, and compare them with the output from the original LLaVA \cite{liu2023llava}. Our approach is able to greatly reduce the amount of \textcolor{brightmaroon}{hallucinations and invalid observations} while increasing the amount of \textcolor{dartmouthgreen}{detailed visual descriptions}. Furthermore, the model retains the \textcolor{denim}{plausible intuitive reasoning} capabilities of the underlying LLM.}
    \label{fig:qualitative}
    \vspace{-15pt}
\end{figure}

\vspace{-10pt}
\subsection{Qualitative Results}

In Figure \ref{fig:qualitative}, we show qualitative results of our technique in action, focusing on a comparative analysis between our model and its starting point (LLaVA). This comparison is particularly relevant as our model undergoes fine-tuning based on the LLaVA model. By juxtaposing these two models, we highlight the enhancements and advancements our model has achieved, offering valuable insights into the efficacy and impact of our training strategies. More results are provided in the Supplementary Materials.

\vspace{-10pt}
\subsection{Fine-grained vs Holistic Evaluation by Reward Model}
\label{sec:ablation}


\begin{wraptable}[10]{r}{0.4\textwidth}
    \vspace{-15pt}
    \centering
    \begin{tabular}{lc}
    \toprule
         Reward model design & MME\\
        \midrule
        none & 960.1 \\
        holistic-based & 1027.4 \\
        fine-grained-based & \textbf{1309.3} \\
        \bottomrule
    \end{tabular}
    \captionof{table}{\textbf{Fine-grained vs holistic evaluation by reward model.}}
    \label{tab:rm_ablation}
\end{wraptable}


In this section, we compare the fine-grained analysis of sentences by the reward model with the more commonly used holistic evaluation of complete responses. This assessment is carried out in the context of reward modeling with human feedback. For the holistic reward model, we retrained a reward model using a singular, unified reward for each description. Specifically, holistic reward is determined based on the content of the entire description, which is negative if the description contains at least one erroneous sentence, and positive otherwise.
As shown in Table~\ref{tab:rm_ablation}, the fine-grain-based approach demonstrates better performance than the holistic approach, which shows that denser and more informative signals allow the reward model to better discern the quality of the generated text by creating more direct links between the visual features in the image and the sentences describing them.

\section{Conclusion, Limitations, and Future Work}
\label{sec:conclusion}

Our work takes a step toward improving visual grounding capabilities of LVLMs to reduce hallucination, errors in relational reasoning, counting, and so on. To this end, we developed a framework named \textbf{ViGoR} to combine fine-grained reward modeling of human preferences with powerful existing open-set visual perception models to efficiently improve LVLMs in these aspects. To validate our approach, we collect the first large dataset with fine-grained human annotation feedback for the LLaVA. We show that ViGoR significantly improves the LLaVA model's ability to generate accurate and relevant text given input images, while preserving its ability to creatively and intuitively reason about the scene.

We note that despite its successes, ViGoR nonetheless exhibits several limitations. The automated component of our reward generation relies heavily on the capabilities of the perception model (i.e. GroundingDINO). As such, current version is limited to objects suitable for detectors while not being applicable to \textit{stuff} regions, attributes, or layouts. Furthermore, the human evaluation data used for training the reward model is specific to a particular LVLM's architecture and checkpoint with which the evaluated responses are generated. For LVLMs with significantly different output or failure modes, additional human preference data may need to be collected, thus incurring further cost. 



We hope to address the aforementioned limitations in future work. As well, we plan to apply RLHF training with our ViGoR framework, which may offer further improvements over our training based on rejection sampling. As well, we anticipate that explicitly linking visual entities with associated phrases in the generated text can further improve visual grounding.  



%
%
\bibliographystyle{splncs04}
\bibliography{main}
\end{document}